\documentclass[review]{elsarticle}

\usepackage{hyperref}
\usepackage{amsmath}
\usepackage{verbatim}
\usepackage{bm}
\usepackage{float}
\usepackage{xcolor}
\usepackage{subcaption}
\floatstyle{plaintop}
\restylefloat{table}
\journal{Signal Processing: Image Communication}

\begin{document}

\begin{frontmatter}

\title{Correlation Net : spatiotemporal multimodal deep learning for action recognition}

\author[mymainaddress,mythirdaddress]{Novanto Yudistira\corref{mycorrespondingauthor}}
\cortext[mycorrespondingauthor]{Corresponding author}
\ead{cbasemaster@gmail.com}
\ead{yudistira@ub.ac.id}
\author[mysecondaddress]{Takio Kurita}
\ead{tkurita@hiroshima-u.ac.jp}
%%\address{Radarweg 29, Amsterdam}
%%\fntext[myfootnote]{Since 1880.}

%%\author[mymainaddress,mysecondaryaddress]{Elsevier Inc}
%%\ead[url]{www.elsevier.com}

%%\author[mysecondaryaddress]{Global Customer Service\corref{mycorrespondingauthor}}
%%\ead{www.elsevier.com}
%%\cortext[mycorrespondingauthor]{Corresponding author}
%%\ead{support@elsevier.com}

\address[mymainaddress]{School of Informatics and Data Science, Hiroshima University, Higashi Hiroshima, 739-8521, Japan }

\address[mysecondaddress]{Department of Information Engineering, Hiroshima University, Higashi Hiroshima, 739-8521, Japan}

\address[mythirdaddress]{Teknik Informatika, Fakultas Ilmu Komputer, Universitas Brawijaya, Malang, 65145, Indonesia}

\begin{abstract}
This paper describes a network that captures multimodal correlations over arbitrary timestamps. The proposed scheme operates as a complementary, extended network over a multimodal convolutional neural network (CNN). Spatial and temporal streams are required for action recognition by a deep CNN, but overfitting reduction and fusing these two streams remain open problems. The existing fusion approach averages the two streams. Here we propose a correlation network with a Shannon fusion for learning a pre-trained CNN. A Long-range video may consist of spatiotemporal correlations over arbitrary times, which can be captured by forming the correlation network from  simple fully connected layers. This approach was found to complement the existing network fusion methods. The importance of multimodal correlation is validated in comparison experiments on the UCF-101 and HMDB-51 datasets. The multimodal correlation enhanced the accuracy of the video recognition results.
\end{abstract}


\end{frontmatter}

%\linenumbers

\section{Introduction}
Video recognition, particularly the recognition of human actions, has progressed from handcrafted to deep learning, which extract the necessary rich spatiotemporal information. The established handcrafted features include dense trajectories \cite{wang2013dense} and their descriptors of Histogram of Flows (HOF), Histogram of Gradients (HOG), Motion Boundary Descriptors (MBH), an improved version of dense trajectories \cite{wang2013action},  action bank that apply steerable filters of spatio temporal space \cite{sadanand2012action}, MOFAP \cite{wang2016mofap} and cross correlation \cite{matsukawa2010action}. Following the rise of deep CNNs, state-of-the-art methods have gradually
come to use deep-learned features because of their scalability and richness of information. However, if there are insufficient training data, deep CNNs give equivalent performance to handcrafted features \cite{karpathy2014large}. In this case, one possibility is to use transfer learning from a bigger dataset followed by fine tuning.

CNNs have been intensively applied to many computer vision tasks, particularly action recognition since it has significantly increased the image classification accuracy on ImageNet challenge. The challenging part, however, is to fuse information from different sources into a combined perception. Recently, action recognition techniques have employed spatial and motion information, which complement one another. Research on information fusion has integrated statistical learning with deep learning fusion schemes for pattern recognition applications. The baseline recognition method over spatiotemporal domains is average pooling, as used by Simonyan et al. \cite{simonyan2014two} for two-stream network and Feichtenhofer  et al. \cite{feichtenhofer2016convolutional} for two-stream network fusion. However, the problem of overfitting
means that there is still a gap between the training and testing datasets. Simple fusions such as sum, average, or multiply can potentially lead to lose the relationship information of both streams, which (if present) would increase generalization of the approach. Moreover, the usual two-stream network adopts a frame-wise training scheme that cannot infer the long-range video classification. Yudistira et al. \cite{yudistira2017gated} proposed a softmax gating mechanism as an additional network for handling stream selection. However, this requires the gating stream to be tuned, which is computationally expensive if the gating stream is also a deep network. If the number of modalities is high, such gating network will be advantageous. However, if there are only two or three modes, it is better to apply a simple network. Instead of using weighted gating scheme, we explore class relationship correlation between streams. The advantages of latter method are capturing long range information and its streams correlation. Recently, fusion based on an independent stream or convolutional stack has been studied, but the associated correlation information has not yet been investigated. If we contemplate with neural code of brain \cite{amari2006correlation}, independent model will cause loss information in stochastic decoding. It relates to deep learning since our optimization is stochastic gradient descent (SGD) by using statistical estimation to minimize objective function.

While the output of CNN using logistic regression and softmax crossentropy are basically probability in nature, however there is still no consideration of correlation between two output modalities. The further discussion and intensive experiment of using correlation network is direction of our investigation. There are several methods of correlation from multi source such as joint probability using element-wise product \cite{zhang2017joint} or concatenation \cite{he2017neural} for task of collaborative filtering. However, it captures less correlation between sources compared to outer correlation \cite{he2018outer}. The latter method captures rich interaction map from two embedding inputs. It is then fed into CNN and increase its previous methods.

For the case of video recognition, long-range temporal information should be considered to obtain better perception. Limin et al. \cite{wang2016temporal} extended the two-stream approach by providing a segmented training scheme. The temporal structure of this scheme improves performance compared with the usual snippet sampling \cite{simonyan2014two}. However, the aggregation process can result in some information loss. A correlation network has the potential to identify this information loss on
a frame-by-frame basis over arbitrary timestamps for the entire video. Several methods have been proposed to capture temporal information on CNN such as \cite{ng2015beyond}, \cite{varol2017long}, \cite{husain2016action}.
However, these are mainly based on dense sampling and a predefined temporal range. Our proposed method has the potential to provide complementary information for multi-modal networks. 

In past research \cite{gers1999learning}, the long-range temporal information has been captured by long short term memory (LSTM). It was introduced to improve Recurrent Neural Network (RNN) \cite{gers1999learning}. In action recognition, such approach was done by \cite{ma2019ts} using BN-Inception backbone. The higher accuracy of LSTM than averaging highlights the importance of including sequential information. This approach, however, is resource-intensive task requiring addition, multiplication, and forget branch to train the paths from previous cells to current one that form complicated network. We propose simpler training procedure that performs comparably to LSTM, and even outperforms LSTM on some datasets.

Motivated by aforementioned problems, we make the following contributions to video recognition research:

\begin{enumerate}
\item Propose a correlation training model that captures spatiotemporal
correlation on a frame-by-frame basis without
time correspondence.
\item Introduce Shannon fusion to select features based on distribution entropy.
\item When applied to a temporally segmented network, the proposed method is shown to provide complementary information for long-range video recognition.
\end{enumerate}

The remainder of our paper is organized as follows. Section \ref{pre} introduces correlated and independent
data. Section \ref{arch} explains the proposed correlation network (Corrnet) architecture. Section \ref{loss} defines loss function and optimization in this work. Then, section \ref{str} describes the training and testing strategies for Corrnet. Section \ref{exp} and section \ref{results} experimentally setup and validate our method on the UCF-101 and HMDB-51 datasets. Drawbacks and advantages of our approach are discussed in Section \ref{discussion}, and Section \ref{conclusion} concludes the paper.

\begin{figure}

\centering
\includegraphics[width=0.8\textwidth]{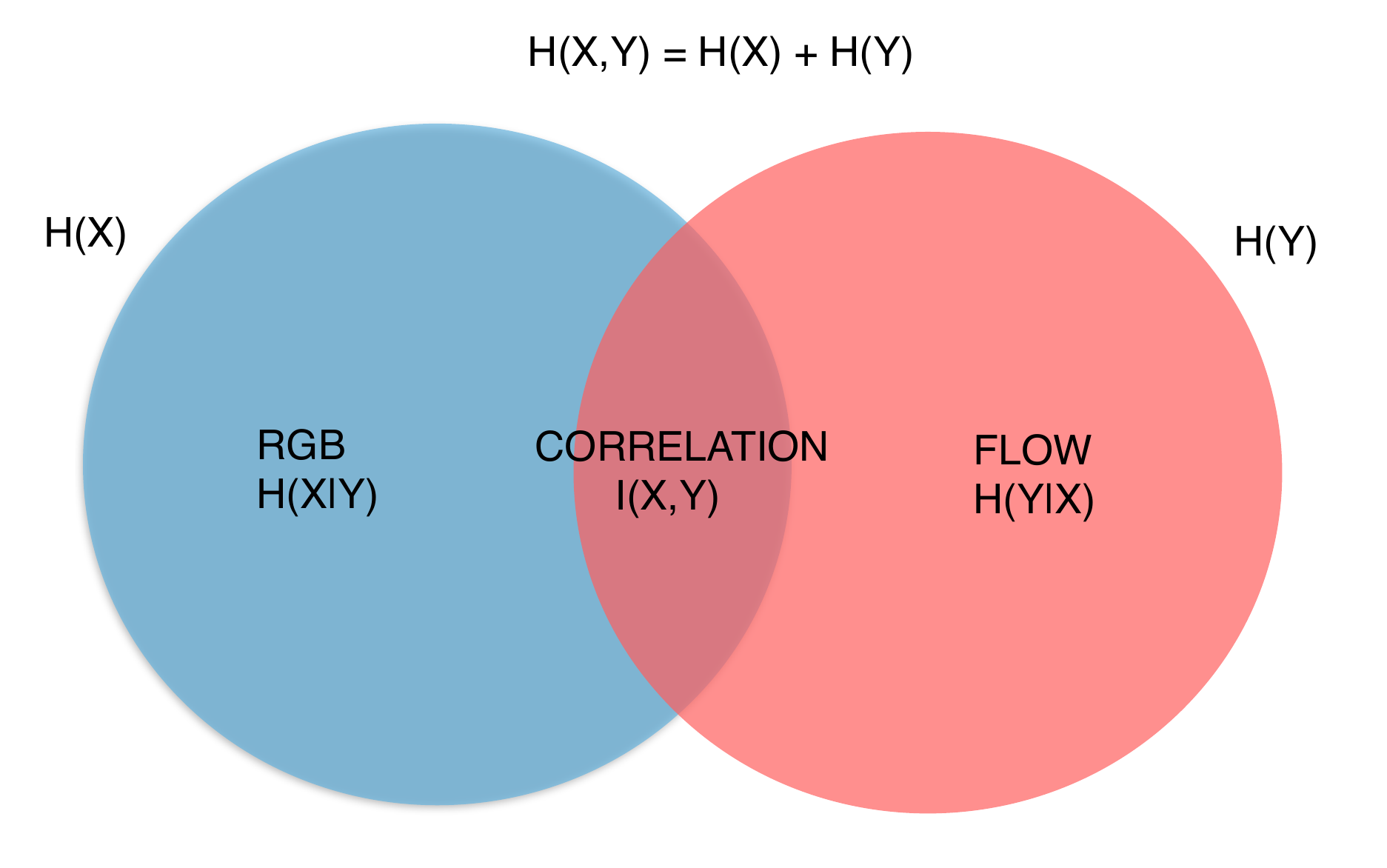}
\centering
\caption{\label{fig:venn}Venn diagaram of independent and correlation stream.}
\end{figure}

\section{Correlated and Independent Data}
\label{pre}
Any given data distribution contains independent and correlated data. The correlated data are the overlapping data of two distributions. These data contain rich information for the deep learning fusion. Action recognition also applies these terminologies, as it recognizes the complementary of multimodal information such as RGB and optical flows. Typical averaging fusion averages only the independent streams without considering their correlated information. Moreover, how the independent streams are correlated would reveal the semantic relationships between classes. Based on information theory \cite{stone2015information}, Figure \ref{fig:venn} shows two individual entropies (RGB and flow streams) and their mutual information (MI) (correlation: dark red area). The MI is considered to represent the mutual dependence between the variables. The variable information is usually measured by the Shannon entropy, which defines the information amount in a given random variable. Entropy is maximized when all members in the variable space are uquiprobable (have equal probability), and minimized when all variable members have unequal probabilities. The entropy of the RGB, flow, and the combined variables are denoted as $H(X)$, $H(Y)$, and $H(X,Y)$, respectively. $H(X)$ and $H(Y)$ are independent if $H(X,Y) = H(X)+H(Y)$. The joint entropy $H(X,Y)$ of two discrete variables is the MI or general version of correlation. Meanwhile, two variables are independent if $H(X|Y) = H(X)$ or $H(Y|X) = H(Y)$. We aim to capture the correlation information and (as an extended data output) how the information is correlated. We then
investigate whether the new information increases the prediction performance on various action datasets.

\section{CNN architecture with correlation network}
\label{arch}

Consider an image sequence of $I = (i_0,i_1,....,i_{t_1})$ and flow sequence of $F = (f_0,f_1,....,f_{t_2})$  where $t_1$ is the number of images over time and $t_2$  is the number of flows over time. Note that each image $i$ contains 3 channels of RGB and $f$ contains 10 consecutive flow field channels. For each iteration $i \in I$ and $f \in F$ are randomly selected. Both are fed into  $S_i(i;W_i)$ (spatial stream) and $S_f(f;W_f)$ (temporal stream) with model parameters of $W_i$ and $W_f$, respectively.
Our architecture is based on two expert streams and one correlation stream. The correlation stream acts as a CNN that can find pattern based on autocorrelation between the two vector outputs. The input for each stream is an arbitrary frame such that, in every iteration, we obtain a random combination of output vectors within the video sequence. This acts as an additional training besides the independently trained spatial and temporal streams. The output of each stream is commonly represented by the class prediction after smoothing with softmax cross-entropy. The
spatial and temporal streams use the BN-Inception network with
batch normalization and weighting, as described by Wang et al. \cite{wang2016temporal}.

\begin{figure}
\centering
\includegraphics[width=1\textwidth]{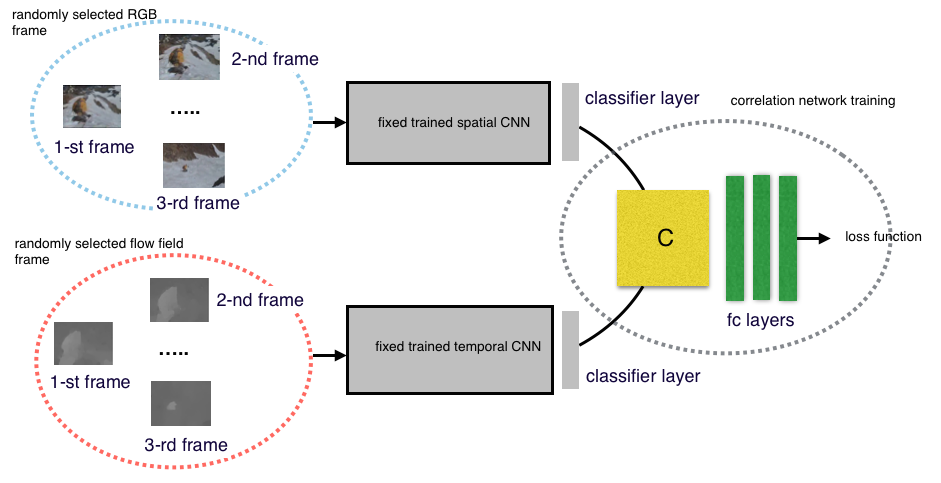}
\centering
\caption{\label{fig:Corrnettrain}Architecture of correlation network which trains output layer of both trained streams.}
\end{figure}

As shown in Figure \ref{fig:Corrnettrain}, the outputs of the two streams are combined into a correlation map. Two-stream architectures can be based on per-frame training or long-range temporal training. In the latter type, long-range video frames are segmented into $K$ parts and the loss function is calculated by summing output of frames from respective segment. This was proposed to handle long range recognition, especially for optical flows, because there is different generalization between one frame and average of all frames.

\begin{figure}
\centering
\includegraphics[width=1\textwidth]{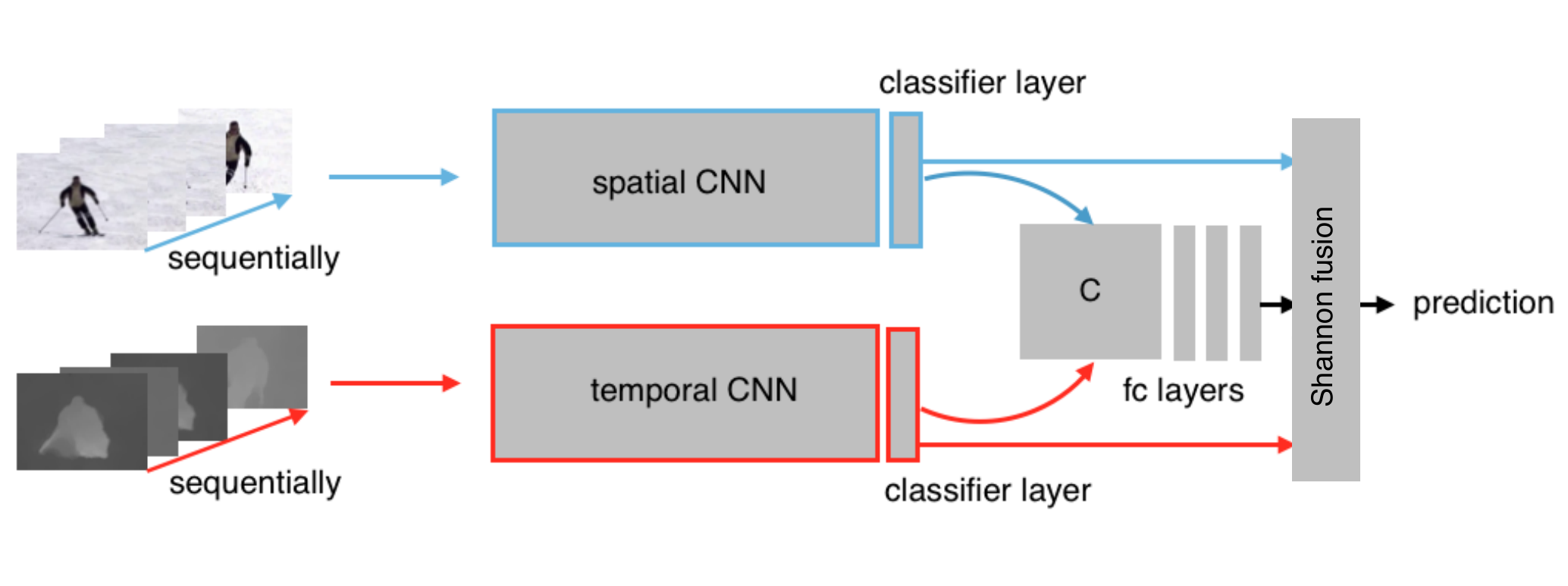}
\centering
\caption{\label{fig:Corrnettest}Testing architecture of correlation network which use both two streams output information as final prediction.}
\end{figure}

The correlation $C$ between the output streams (Figure \ref{fig:Corrnettrain}) is calculated as the outer product between the spatial ($\bm{u}$) and the temporal ($\bm{v}$) output vectors of the spatial stream and the temporal stream, respectively:

\begin{equation}
C = \bm{u} \otimes \bm{v} = \bm{u} \bm{v}^T
\end{equation}

 Note that both $\bm{u}$ and $\bm{v}$ are class scores which may be negative or positive. Suppose $\bm{u} = \begin{bmatrix}
           u_{0} \\
           u_{1} \\
           \vdots \\
           u_{n-1}
         \end{bmatrix}$ and $\bm{v} = \begin{bmatrix}
           v_{0} \\
           v_{1} \\
           \vdots \\
           v_{m-1}
         \end{bmatrix}
$ are vector of spatial and motion with element index of $1,2,..n$ and $1,2,..m$, respectively. Note that both $n$ and $m$ are positive integer. The outer product of both vectors should be 2-dimensional of size of $n \times m $ (eq. 2) in which indexed by two subscripts of $n$ and $m$. The idea of using outer product is to express rich semantic relationships. It is not only subsumes the interaction signal of its diagonal elements which equals to inner product,
but also includes all other pairwise correlations.
%Outer product has dimension of equal to product of two factors. 
After L2 normalization of each row, C becomes: 

\begin{equation}
\hat{C}=
  \begin{bmatrix}
    \frac{u_{0} v_{0}}{\lVert \bm{\beta}_{0} \rVert} & \frac{u_{0} v_{1}}{\lVert \bm{\beta}_{0} \rVert} & \dots & \frac{u_{0} v_{M}}{\lVert \bm{\beta}_{0} \rVert}  \\
    \frac{u_{1} v_{0}}{\lVert \bm{\beta}_{1} \rVert} & \frac{u_{1} v_{1}}{\lVert \bm{\beta}_{1} \rVert} &  & \frac{u_{1} v_{m}}{\lVert \bm{\beta}_{1} \rVert} \\
    \vdots &  & \ddots & \vdots \\
    \frac{u_{n} v_{0}}{\lVert \bm{\beta}_{n} \rVert} & \frac{u_{n} v_{1}}{\lVert \bm{\beta}_{n} \rVert} & \dots & \frac{u_{n} v_{m}}{\lVert \bm{\beta}_{n} \rVert}
  \end{bmatrix}
\end{equation}
{where $\bm{\beta}_n$, $n$, and $m$ are the $n$-th row of $C$, the number of elements in the vector $\bm{u}$, and the number of elements in the vector $\bm{v}$, respectively. This matrix contains the interaction map of class relationship, represented not only by the correlations between each pair of same-indexed elements from the two vectors but also by combinations of elements with different indices. The semantic relationships among the classes can then be constructed. In this representation, the data types are real-valued numbers, that can be positive or negative. 

\begin{eqnarray}
(\bm{u} * \bm{v})[l,k] = \sum_{i=0}^{m-1}\sum_{j=0}^{n-1} \bm{u}[i,j]\bm{v}[i+l,j+k]
\end{eqnarray}
\newline

In eq. (3), $*$ denotes cross correlation. $m$ and $n$ are numbers of elements in vector $\bm{u}$ and $\bm{v}$, respectively. It measures whether 2-dimensional cross correlation matrix is either positively correlated ($\gg0$), negatively correlated ($\ll0$), or not correlated ($\approx0$) given large variation of values inside $n \times m$ matrix. For example, if one element contains negative and the other positive signs or vice versa, their product will contain positives and negatives, in which case the two can be considered as uncorrelated because the sum of its product approaches zero. Intuitively, if two vectors have opposite signs, then the integral of the lag between $j$ and $k$ is small. If both elements are negatives, a negative correlation is implied.} Therefore, there will be high variation of number and sign to infer that both vectors are correlated or not and it is beneficial for network. For the case of element wise multiplication, the product of vector has the size of $m$ or $n$ (same as output of both stream) of which has less variation than outer product.

Tensor $C$ is then flattened and fed to a multilayer perceptron consisting of three fully
% * <mkkavi14@gmail.com> 2018-10-06T07:53:01.603Z:
% 
% > Perceptron c
% change to small letter
% 
% ^.
connected classifier layers: fc1 whose input dimension is 
same as the output of each stream and output dimension is
4096; fc2 with an input and output dimensions are 4096; and fc3 whose input dimension of 4096 and an output dimension equaling the number of classes. The fully connected layers adjust their weights during the training process.

\section{Loss Function}
\label{loss}

To optimize spatial, motion, and correlation stream, we use separate loss and optimizer of stochastic gradient descent (SGD). The total loss function is given by:
%\newline
%\newline
%\newline
\begin{eqnarray}
L = -\sum_{b=1}^B y_b\big(\log(z_b) + \log(p_b)\big)
\end{eqnarray}
\newline
where $y_b$ is the ground truth label corresponding to class b while $z_b$ and $p_b$ are softmax output of Corrnet and two stream in index or class $b$, respectively. We can elaborate $z_b$ and  $p_b$ as: 
\begin{align}
  z_b = \frac{\exp(Z(\bm u,\bm v;W_c)_b)}{\displaystyle \sum_{j=1}^{B} \exp(Z_j)}, \ \ 
  p_b = \frac{\exp(g_b)}{\displaystyle \sum_{j=1}^{B} \exp(g_j)}.
\end{align}
In this expressions, $Z$ is the Corrnet model with parameter $W_c$. $z$ is the Corrnet output, and $B$ is the number of classes. The spatial and temporal outputs are summed as $\bm g=\bm u+\bm v$. If the spatial and the motion streams are fixed, which is the case for CNN streams that have been trained and fixed, the loss function is  $-\sum_{b=1}^B y_b\log(z_b)$. Both are the same in terms of optimization. To update network by backpropagation, we calculate the gradients
of the loss function as:
\begin{align}
\frac{\partial{L}}{\partial{W_c}}&=\frac{\partial{L}}{\partial{Z}}\frac{\partial{Z}}{\partial{W_c}} \\
%\frac{\partial{fc2}}{\partial{fc1}}\frac{\partial{fc1}}{\partial{W_c}}
\frac{\partial{L}}{\partial{W_i}}&=\frac{\partial{L}}
{\partial{g}}\sum_{k=0}^{K}\frac{\partial{g}}{\partial{S_i(T_k)}}\frac{\partial{S_i(T_k)}}{\partial{W_i}} \\
\frac{\partial{L}}{\partial{W_f}}&=\frac{\partial{L}}
{\partial{g}}\sum_{k=0}^{K}\frac{\partial{g}}{\partial{S_f(T_k)}}\frac{\partial{S_f(T_k)}}{\partial{W_f}}
\end{align}

As indicated in eq. (5), Corrnet is optimized independently on the spatial and temporal streams. Moreover, through the fully connected layers with parameters $W_c$, the correlation structure is learned through backpropagation. Eq. (6) and (7) describe the backpropagation flow with respect to the spatial parameters ($W_i$) and temporal parameters ($W_f$), respectively. Both gradients adopt segmental consensus in temporal segment network (TSN) with $K$ segments. $T_k$ is the selected frame in segment $k$.

\begin{figure}[ht!]
\centering
\begin{subfigure}{.5\textwidth}
  \centering
  \includegraphics[width=1.\linewidth]{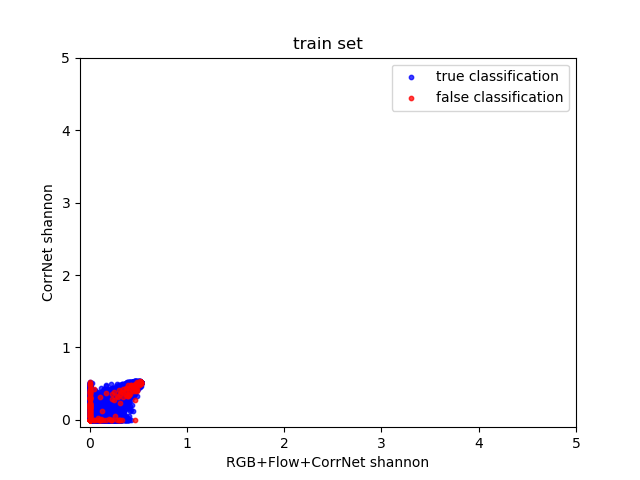}
  \caption{}
  \label{fig:shannon_train}
\end{subfigure}%
\begin{subfigure}{.5\textwidth}
  \centering
  \includegraphics[width=1.\linewidth]{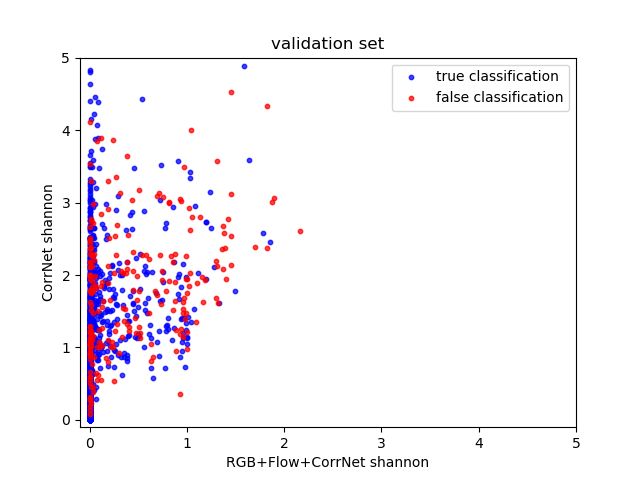}
  \caption{}
  \label{fig:shannon_val}
\end{subfigure}
\caption{Shannon fusion of Correlation Net and total fusion on (a) training and (b) validation set.}
\label{fig:shannon_all}
\end{figure}

\section{Training and testing strategies for the correlation network}
\label{str}

A two-stream network architecture was selected because TSNs have proven successful in previous experiment \cite{wang2016temporal}. The architecture incorporates a BN-Inception CNN, which effectively balances the accuracy–--speed tradeoff. The learned weights are then transferred and fixed, meaning that no updates are processed while training the two-stream network. As the dimension of correlation tensor $C$ is $dim(\bm{u})\times dim(\bm{v})$, the correlation architecture can be a CNN or a multilayer perceptron. The features for testing are selected by Shannon fusion, which identifies the dominant element by the entropy computation. The Shannon entropy is defined as:

\begin{equation}
SE = -\sum_{i=0}^{N-1} (\bm\hat{q_i}) \log_2 (\hat{q_i})
\end{equation}

where $\hat{q_i}$ is $i$-th element of a normalized vector $\bm{\hat{q}}$. Corrnet is more easily distracted by an input vector with sparse entropy than a vector with a few dominant elements. It is trained using input from converged source streams (RGB and flow) of which indicate low Shannon entropy (high confidence). Output of trained Corrnet is also converged and thus fusion of all outputs (RGB, flow, and Corrnet) also have low entropy (Fig \ref{fig:shannon_train}). Different from train set, output of source streams (RGB and flow) in validation set is somewhat in lower confidence compared to train set. As shown in Fig \ref{fig:shannon_val}, outputs of validation set have either low or moderate entropy. Therefore, Shannon fusion is selection method that if eq. (10) is met, the Corrnet output is excluded from the fusion.  

\begin{equation}
SE(softmax(\underbrace{Z(\bm u, \bm v)}_\text{Corrnet output}+\underbrace{\bm u+\bm v}_\text{2 stream output}\underbrace{-\min(Z(\bm u, \bm v))-\min(\bm u+\bm v)}_\text{normalization})) \geq th
\end{equation}

 Each Corrnet output and 2 stream output are subtracted by its minimum element; therefore the lowest value of output vector becomes zero. Threshold of $th$ is used as limit to decide whether Corrnet is included in fusion or not. Threshold of $th$ is searched heuristically within certain of range (min=0,max=$log{N}$ ). Small training simulation by splitting train set into sub-train and sub-val set can be held to find value of $th$. In all experiments of using BN-Inception with TSN backbone, we found that $th$ of 1.0 is achieved after such cross-validation procedure was done. For experiment done by using I3D backbone, it is found that $th$ of 2.0 is the best threshold. The final fusion of testing scheme is $softmax(Z(\bm u, \bm v)+\bm u+\bm v)$ or $softmax(Z(\bm u, \bm v)+0.5(\bm u+\bm v))$.

The performances of TSN and a two-stream network with and without Corrnet were assessed on equally spaced 24 RGB images and optical flow stacks to spatial and temporal nets, respectively. We divide the 24 sampled frames into K equal segments. In our experiments, we set $K=3$. For TSN training, in each sampled frame, we obtained 10 inputs by cropping and horizontally flipping the four corners and the center region.

\section{Experiment}
\label{exp}

Experiments were conducted on UCF-101 \cite{soomro2012ucf101}, HMDB-51 \cite{kuehne2011hmdb} and Charades dataset \cite{sigurdsson2016hollywood}. The UCF101 dataset contains 13320 videos divided into 101, whereas HMDB51 contains 6766 videos in 51 classes. All videos were collected from Youtube and have many degree of freedom, which complicate the task of video or action classification. Both datasets are split three ways with different combination of training and testing data. To ensure a fair comparison with previous methods, we trained and tested our scheme on the original dataset. The results of split 1 and the average results all over splits were reported for each dataset. The Charades dataset contains 9848 videos of daily activities, which have been multi-labeled by Amazon Mechanical Turk. The videos were recorded by 267 different actors, and are divided into 157 classes with 66500 temporal annotations. To handle the multi-label nature of this dataset, we applied the  sigmoid activation, binary cross-entropy as the loss function, and the mean average precision (mAP) for evaluation.

The flow modality was the optical flow extracted by the total variation algorithm with data fidelity measured in the L1 norm \cite{zach2007duality}. The flow was generated by the OpenCV framework. The magnitude and angular in the flow was linearly transformed into RGB images in range of 0-255 using a linear transformation. Our correlation network in the deep learning framework, we used the Chainer \cite{tokui2015chainer} for TSN with Imagenet pre-train, Pytorch \cite{paszke2017automatic} for TSN with Kinetics pre-train, and Tensor Flow \cite{abadi2016tensorflow} for I3D with Kinetics pre-train to train and test our correlation network. The two-stream network were TSN, I3D and two-stream network \cite{wang2015towards} with Caffe-trained weights \cite{jia2014caffe}. The TSN and two-stream networks were implemented in the BN-Inception and the two-stream VGGNet-16 networks, respectively. TSN was trained using temporal segment network essentially two-stream CNN network. The training was segmental to capture long-range video actions. This network previously achieved state of the art results on UCF101 and HMBD51.  The correlation network was optimized by stochastic gradient descent (SGD). Mini-batches of 8 were trained through 200 epochs, with momentum of 0.9 and a learning rate of 0.001. For the implementation of using I3D, each sampled clip contains 64 frames for both RGB and optical flows. In testing, the output of Corrnet from I3D streams is weighted by 0.01 to make it equal to the output of spatial and motion stream.  

As the HMDB-51 dataset has fewer training data than UCF101, a transfer learning approach was used in which network was trained on UCF-101 split 1, and then applied to all splits of HMDB-51. Because the HMDB-51 output has 51 classes whereas UCF 101 has 101 classes, the number of trained weights in the first layer of the correlation network was increased to 101 by a tiling strategy (fc1). The network was then fine-tuned for training on the HMDB-51 dataset. This procedure was used in order to speed up learning even though there is no evidence about increasing performance.

As the RGB and flow in the Charades
dataset, we used the I3D stream trained on the Kinetics dataset. Again, the correlation network was optimized by SGD. Mini-batches of 4 runs were trained through 64000 steps, with a momentum of 0.9 and a learning rate of 0.001. 

We evaluated two types of late fusion in correlation net: the averaged and non-averaged final fusion of both streams. We compared the performances of TSN and the two-stream network without correlation net and with additional correlation network. The TSN is pre-trained on Imagenet or Kinetics. The latter pre-trained dataset contains approximatey 650000 video clips \cite{carreira2017quo}. The TSN was trained on three segments. 

\section{Results}
\label{results}
\begin{table}[H]
  \centering
  \begin{tabular}{|c|c|c|}
  \hline
     & accuracy\\
    \hline
    Two stream & 89.4  \\
    \hline
    avg(Two stream) + Corrnet & \bf{89.7}  \\
    \hline
    Two stream + Corrnet & \bf{89.9}   \\
    \hline
    TSN (Imagenet pre-trained) & 93.5   \\
    \hline
    avg(TSN) + Corrnet (Imagenet pre-trained) & \bf{94.2}\\
    \hline
    TSN + Corrnet (Imagenet pre-trained + Shannon fusion) & \bf{94.2}   \\
    \hline
    TSN (Kinetics pre-trained) & 94.3   \\
    \hline
    avg(TSN) + Corrnet (Kinetics pre-trained) & \bf{94.6}   \\
    \hline
    TSN + Corrnet (Kinetics pre-trained + Shannon fusion) & 94.5   \\
    \hline
    I3D (Kinetics pre-trained) & 97.8   \\
    \hline
    I3D + Corrnet (Kinetics pre-trained) & \bf{98.1}   \\
    \hline
    I3D + Corrnet (Kinetics pre-trained + Shannon fusion) & \bf{98.1}   \\
    \hline
  \end{tabular}
  \caption{Accuracy performance of TSN and the two-stream methods on UCF-101 split 1}
  \label{tab:1}
\end{table}

As clarified in Table \ref{tab:1}, the method using Corrnet improved the performance on UCF-101 Split 1. The results of TSN and the two-stream network were based on the weights transferred from their original Caffe implementation. TSN performance was improved by the correlation network, highlighting the importance of additional spatiotemporal correlation. On Split 1, the correlation network improved the TSN and the original two-stream network by 0.7\% and 4.8\%, respectively. When using Kinetics pre-trained, TSN performance is better than Imagenet pre-train by 0.8\% as well as its fusion with Corrnet by 0.3\%. When I3D backbone with Kinetics pre-trained is used, better performance is achieved by 97.8\%. With correlation network, the performance is improved by 0.3\% becoming 98.1\%.

\begin{table}[H]
  \centering
  \begin{tabular}{|c|c|c|c|c|}
  \hline
    & split 1 & split 2 & split 3 & average\\
    \hline
    spatial &  85.9 & 84.9 & 84 & 84.9\\
    \hline
    motion &87.9 & 90.3 & 91 & 89.7\\
    \hline
    Corrnet & 88.3 & 87.6 & 87.9 & 87.9\\
    \hline
    s+m & 93.5 & 94.5 & 94 & 94\\
    \hline
    avg(s,m)+Corrnet & \bf{94.2} & \bf{94.6} & \bf{94.1} & \bf{94.3} \\
    \hline
    s+m+Corrnet (Shannon fusion) & \bf{94.2} & \bf{94.7} & \bf{94.2} & \bf{94.4} \\
    \hline
  \end{tabular}
  \caption{UCF-101 pre-trained on Imagenet all splits accuracy}
  \label{tab:2}
\end{table}

On UCF-101 splits 2 and 3
pre-trained by Imagenet (Table\ref{tab:2}), the correlation network improved the TSN performance  0.2\%. The overall average of TSN with Corrnet using Shannon fusion was 94.4\%, further confirming that Corrnet achieved higher recognition fidelity than TSN and the two-stream network.

\begin{table}[H]
  \centering
  \begin{tabular}{|c|c|c|c|c|}
  \hline
    & split 1 & split 2 & split 3 & average\\
    \hline
    spatial &  54.3 & 50.2 & 50.4& 51.6\\
    \hline
    motion & 62.3 & 63.5 & 64.2 & 63.3\\
    \hline
    Corrnet & 66.6 & 65.8 &65.5&66\\
    \hline
    s+m & 69.9 & 67.1 & 67.1&68\\
    \hline
    avg(s,m)+Corrnet& \bf{70.6} & \bf{67.9} & \bf{67.8} & \bf{68.8}\\
    \hline
    s+m+Corrnet (Shannon fusion)& \bf{70.6} & \bf{67.9} & \bf{68.1} &\bf{69}\\
    \hline
  \end{tabular}
  \caption{HMDB-51 pre-trained on Imagenet all splits accuracy}
  \label{tab:3}
\end{table}

Table \ref{tab:3} reports the accuracy result on HMDB-51 pre-trained by Imagenet. On Splits 1, 2 and 3, the Corrnet improved the TSN performance by 0.7\%,
0.8\% and 1.0\%, respectively. The overall accuracies of the average and motion-plus-spatial versions were 68.8\% and 69\% respectively when the
correlation network was included.

\begin{table}[H]
  \centering
  \begin{tabular}{|c|c|c|}
  \hline
 & Val mAP \\
    \hline
    I3D (RGB) &  35.0\\
    \hline
    I3D (Flow) & 10.3\\
    \hline
    I3D (RGB) + Corrnet (RGB) & \bf{35.2} \\
    \hline
    I3D (RGB + Flow) & 32.8 \\
    \hline
    I3D (RGB + Flow) + Corrnet & 32.8 \\
     \hline
  \end{tabular}
  \caption{Charades dataset accuracy}
  \label{tab:4}
\end{table}

Table \ref{tab:4} reports the results on the Charades dataset using Corrnet trained on outputs of pair-wise RGB fused with output of RGB stream, there is improvement of 0.2\% over RGB stream, while on Corrnet trained with output of RGB and flow, fusion with RGB and flow does not make any improvement since the performance of flow stream is lower than RGB by large margin of 24.7\%.

Next, we compared the performance of Corrnet and other methods of fusing both modalities. The selected methods were averaging (ava), maximum (max), multiply \cite{simonyan2014two}\cite{wang2017scene}, and canonical correlation analysis (CCA) \cite{miao2017multimodal} fusion method using the same BN-Inception stream backbone. For multi layer perceptron, we use same network structure of 3 layers of fc layer as described in section \ref{arch}.

\begin{table}[H]
\scalebox{0.85}{
  \centering
  \begin{tabular}{|c|c|c|c|c|c|c|}
  \hline
    Fusion methods & classifier & ava & max & multiply & CCA & Corrnet \\
    \hline
    UCF101  & class score & 93.5 & 91.3 & 93.3 & 93.3 & - \\
    & multi layer perceptron+s+m & 93.8 & 93.6 & 94.0 & 94.0 & \bf{94.2} \\
    \hline
    HMDB51  & class score & 69.9 & 64.8 & 64.9 & 65.6 & - \\
    & multi layer perceptron+s+m & 70.1 & 69.7 & 68.6 & 68.9 & \bf{70.6} \\
    \hline
  \end{tabular}
  }
  \caption{Comparison to another fusion methods using BN-Inception network on split 1}
  \label{tab:5}
\end{table}

As in table \ref{tab:5}, if vectors generated from operations (ava, max, multiply, and CCA) are trained to multi layer perceptrons and fused with RGB and flow stream, there are improvements over raw class scores on both UCF101 and HMDB51 dataset. Corrnet outperforms trained ava, max, multiply, and CCA by 0.5\%, 0.9\%, 5.1\%, and 3.9\%, respectively. Corrnet outperforms ava, max, multiply, and CCA trained on multilayer perceptron by 0.4\%, 0.6\%, 0.2\%, and 0.2\%, respectively on UCF101 split 1. While on HMDB51 split 1 with regards of fusion with spatial and motion stream, Corrnet outperforms ava, max, multiply, and CCA by 0.2\%, 0.9\%, 2\%, and 1.7\%, respectively.

In Table \ref{tab:6}, we also evaluate the proposed method in comparison with another architecture such as two stream late fusion of Feichtenhofer (2 VGG-M \& 2 VGG-19) \cite{feichtenhofer2016convolutional}, multiplicative way (VGG-19) \cite{park2016combining} and TSN with gating network (2 bn-inception \& 1 VGG-16) \cite{yudistira2017gated}. Independent streams of TSN with Corrnet gives best performance over multiplicative, late fusion of 2 VGG-M, and late fusion of 2 VGG-16 with margin of  5\%, 8.16\%, and 3.48\%, respectively on UCF101 split 1 and same accuracy with gating network on UCF101 split 1 with fewer number of parameters (\textgreater150M of gating network to \textless100M of ours).

\begin{table}[H]
  \centering
  \scalebox{0.8}{
  \begin{tabular}{|c|c|c|}
  \hline
    & UCF-101 & HMDB-51 \\
    \hline
    IDT+FV \cite{wang2013action}  &85.9&57.2 \\
    \hline
    IDT+HSV \cite{peng2016bag} &87.9& 61.1\\
    \hline
    MoFAP \cite{wang2016mofap} &88.3& 61.7\\
    \hline
    TDD+FV \cite{wang2015action} &90.3& 63.2 \\
    \hline
    Two-stream \cite{simonyan2014two} &88&59.4   \\
    \hline
    TSN (imagenet pre-train)\cite{wang2016temporal}&94&68   \\
    \hline
    Two-stream I3D (imagenet pre-train)
    \cite{carreira2017quo}&93.4&66.4\\
    \hline
    TSN LSTM (imagenet pre-train) \cite{ma2019ts}&94.1&69\\
    \hline
    TSN Corrnet (imagenet pre-train) with Shannon fusion (ours) &\bf{94.4}&\bf{69}\\
    \hline
  \end{tabular}
  }
  \caption{Comparison to the state of the art on UCF-101 }
  \label{tab:6}
\end{table}

We also compared proposed Corrnet with state-of-the-art techniques using HMDB51 and UCF101 datasets. The results in Table \ref{tab:6} compare the handcrafted to deep learned features. The comparative methods use Fisher Vector (FV) and Hybrid Supervector (HSV) of improved trajectory (IDT) features \cite{wang2013action}\cite{peng2016bag}, and Multi-Level Motion Features (MoFAP) \cite{wang2016mofap}. The good result is obtained by applying handcrafted of Fisher Vector (FV) encoding on end to end learning of Trajectory-Pooled Deep-Convolutional Descriptors (TDD) \cite{wang2015action}. The full end to end learning of two stream with SVM fusion gives reliable performance, however, still performs below our proposed method. CNN learning using temporal segment strategy of TSN gives better accuracy than previous per frame based two stream \cite{wang2016temporal}. As shown in Table \ref{tab:6}, the results of our TSN and correlation network outperform TSN by 0.4\% and 1\% on UCF-101 and HMDB-51, respectively. We also compare the performance results of the proposed method and LSTM on the UCF101 and HMDB51 since LSTM also capture temporal dynamics of video frames. We report the result on UCF101 and HMDB51 dataset compared with LSTM on average of split 1, 2, and 3. There is improvement of 0.3\% over TSN-LSTM on UCF 101 and same performance on HMDB51.

 The performances of Corrnet fusion and the two-stream late fusion described by Feichtenhofer \cite{feichtenhofer2016convolutional}, a multiplicative approach \cite{park2016combining}, and TSN with a gating network \cite{yudistira2017gated} are compared in Table \ref{tab:7} in terms of accuracy and number of parameters.

\begin{table}[H]

 \scalebox{0.75}{
  \begin{tabular}{|c|c|c|c|c|}
  \hline
    Fusion methods & \# of parameters & spatial & motions & fusion \\
    \hline
    multiplicative \cite{park2016combining} & \textgreater138M (VGG19) & - & - & 89.1\\
    \hline
    Two-Stream (late fusion) \cite{feichtenhofer2016convolutional}& \textless181.42M (2 VGG-M)& 74.2 & 82.3 & 85.94\\
    \hline
    Two-Stream (late fusion) \cite{feichtenhofer2016convolutional}& 257M (2 VGG-16)&  82.6 &  86.3 & 90.62\\
    \hline
    Two-Stream (ReLU5 + FC8) \cite{feichtenhofer2016convolutional}&181.68M (2VGG-M) & -& -& 86.04\\
    \hline
    TSN gating  \cite{yudistira2017gated}& \textgreater150M (2 bn-inception \& 1 VGG-16)& 85.9 & 87.9 & \bf{94.2}\\
    \hline
    TSN Corrnet (ours)  & \textless100M (2 bn-inception \& 1 Corrnet) & 85.9 & 87.9 & \bf{94.2}\\
    \hline
  \end{tabular}
}  
\caption{Comparison to another architectures on split 1}
\label{tab:7}
\end{table}

Table \ref{tab:7} demonstrates that correlating the independent streams of TSN improved the performance on UCF-101 Split 1 dataset. The improvement margin was  3.58\% over late fusion and 5.1\% over the multiplicative and gating network approaches, despite the fewer parameters in our scheme than the earlier schemes.
% * <mkkavi14@gmail.com> 2018-10-06T08:24:06.230Z:
% 
% > Deep-Convolutional Descriptors 
% Please check capital letters?
% 
% ^.
% * <mkkavi14@gmail.com> 2018-10-06T08:23:12.317Z:
% 
% > multi-level motion features (MoFAP) 
% Is is correct? MoFAP
% 
% ^.
% * <mkkavi14@gmail.com> 2018-10-06T08:21:47.810Z:
% 
% > Fisher Vector (FV) and Hybrid Supervector (
% please check capital and small letters
% 
% ^.

\section{Discussion}
\label{discussion}

\subsection{Advantages and drawbacks}

In this section we would like to discuss the drawbacks and advantages of the proposed method. The proposed method of fusion with Corrnet delivers good performance when the accuracy of two streams are balanced. However, if the both two streams produce very high performance, the improvement is small. In that situation, independent streams leaves little room for correlation network to contribute. As shown in Table \ref{tab:8}, Imagenet pre-trained TSN fused with Corrnet gives improvement of 0.5\% compared to Kinetics pre-trained TSN of which only gives improvement of 0.3\%. On HMDB51 split 1 as shown in Table \ref{tab:3}, fusion with Corrnet improves original TSN with improvement of 0.7\%. On contrary, when dealing with severe imbalanced streams, Corrnet gives no contribution to performance. As shown in Table \ref{tab:4} on Charades dataset, the performance of flow stream is very low of 10.3\% compared to RGB of 35\%. In that situation, fusion with Corrnet gives same performance with original of 32.8\%. 

\begin{table}[H]
\scalebox{0.75}{
  \centering
  \begin{tabular}{|c|c|c|c|}
  \hline
    Methods& Backbone&Imagenet pre-train & Kinetics pre-train\\
    \hline
    Corrnet & TSN-BN-Inception&88.3 &93.0   \\
    \hline
    RGB + Flow  &TSN-BN-Inception& 93.5& 94.3   \\
    \hline
    RGB + Flow + Corrnet &TSN-BN-Inception& 94.0&94.6   \\
    \hline
    RGB + Flow + Corrnet (Shannon fusion) &TSN-BN-Inception& 94.2&94.5\\
    \hline
    RGB + Flow  &I3D& -&97.8\\
    \hline
    RGB + Flow + Corrnet &I3D& -&98.1\\
    \hline
    RGB + Flow + Corrnet (Shannon fusion) &I3D& -&98.1\\
    
    \hline
  \end{tabular}
  }
  \caption{UCF-101 pre-trained on Kinetics split 1 accuracy}
  \label{tab:8}
\end{table}

Shannon fusion as method to select whether Corrnet is included in fusion or not as previously defined in section 5 will work best if sum of independent streams (RGB and Flow in this case) is in moderate performance. As shown in Table \ref{tab:8}, Corrnet with Kinetics pre-trained TSN backbone gives 93\% accuracy on UCF101 of which only 1.3\% difference with fusion of RGB and Flow. Whereas, Corrnet with Imagenet pre-trained TSN backbone on UCF101 delivers difference of 5.2\% of which increase of 0.2\% is obtained in contrast with the former which decreasing performance with 0.1\%. The best performance will be achieved using fusion of Corrnet with 2 stream using I3D backbone of 98.1\%. It gains increase of 0.3\% from baseline Kinetics pre-trained I3D. It is comparable with current state of the art results.

\subsection{Independence and correlation}
Given distribution of data, there exist independence and correlated data (Fig. \ref{fig:venn}). The performance of Corrnet is affected by underlying distribution of data. Thus, there is possibility that Corrnet performs better or lower than independent streams fusion, however, it is reasonable to be complementary with existing independent outputs of RGB and Flow. As shown in Table \ref{tab:2}, even though the performances of Corrnet are lower than sum fusion of RGB and flow, its fusion with existing streams increases accuracy on all splits. Same phenomena occur on HMDB51 dataset in Table \ref{tab:3}, which confirms the complementary information of correlation to independent streams.  

The characteristic of our proposed method is based on assumption of correlation between streams. It gives identical performance with gating CNN of \cite{yudistira2017gated} on UCF101 split 1 as shown in Table \ref{tab:7}. We use the same RGB and flow stream network of BN-Inception with TSN scheme and thus have the same performance of spatial and temporal stream. Gating CNN requires additional network (gating stream) functioned as weighting the output of spatial and temporal stream before fusion. Gating stream has to be capable of weighting spatial stream more if spatial features such as shapes or objects are more salient than temporal features (motion), and vice versa. It leads to performance improvement over original TSN. Similarly, our proposed method uses simple network to perform correlation learning before fusion. Our network is simpler indicated by lower number of parameters. Moreover, on HMDB51 split 1, fusion with Corrnet performs better than gating stream which, conclude the beneficial of Corrnet over gating CNN.

\begin{figure}[H]
\centering
\begin{subfigure}{.8\textwidth}
  \centering
  \includegraphics[width=1.\linewidth]{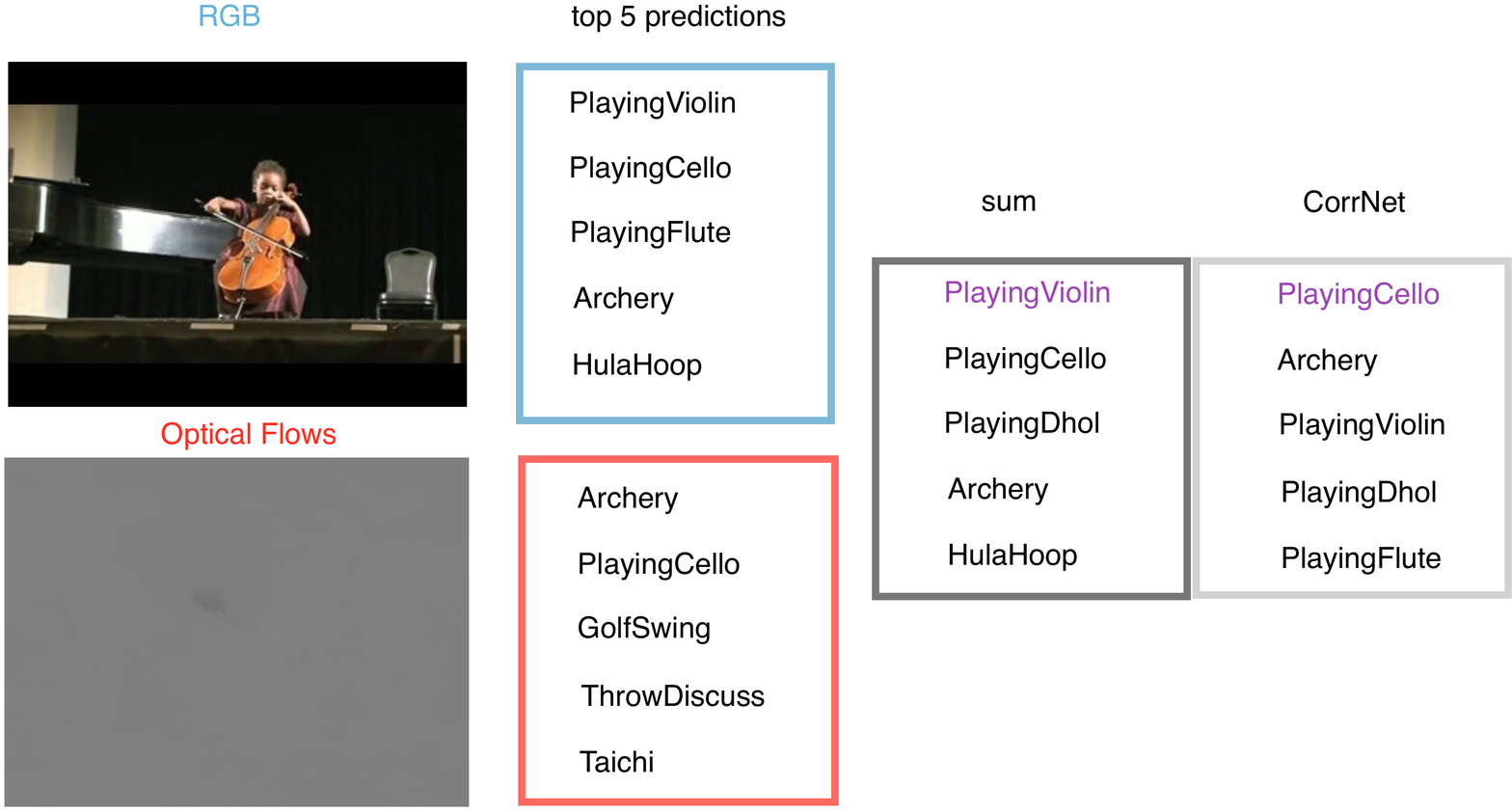}
  \caption{}
  \label{fig:class_score1}
\end{subfigure}\\
\begin{subfigure}{.8\textwidth}
  \centering
  \includegraphics[width=1.\linewidth]{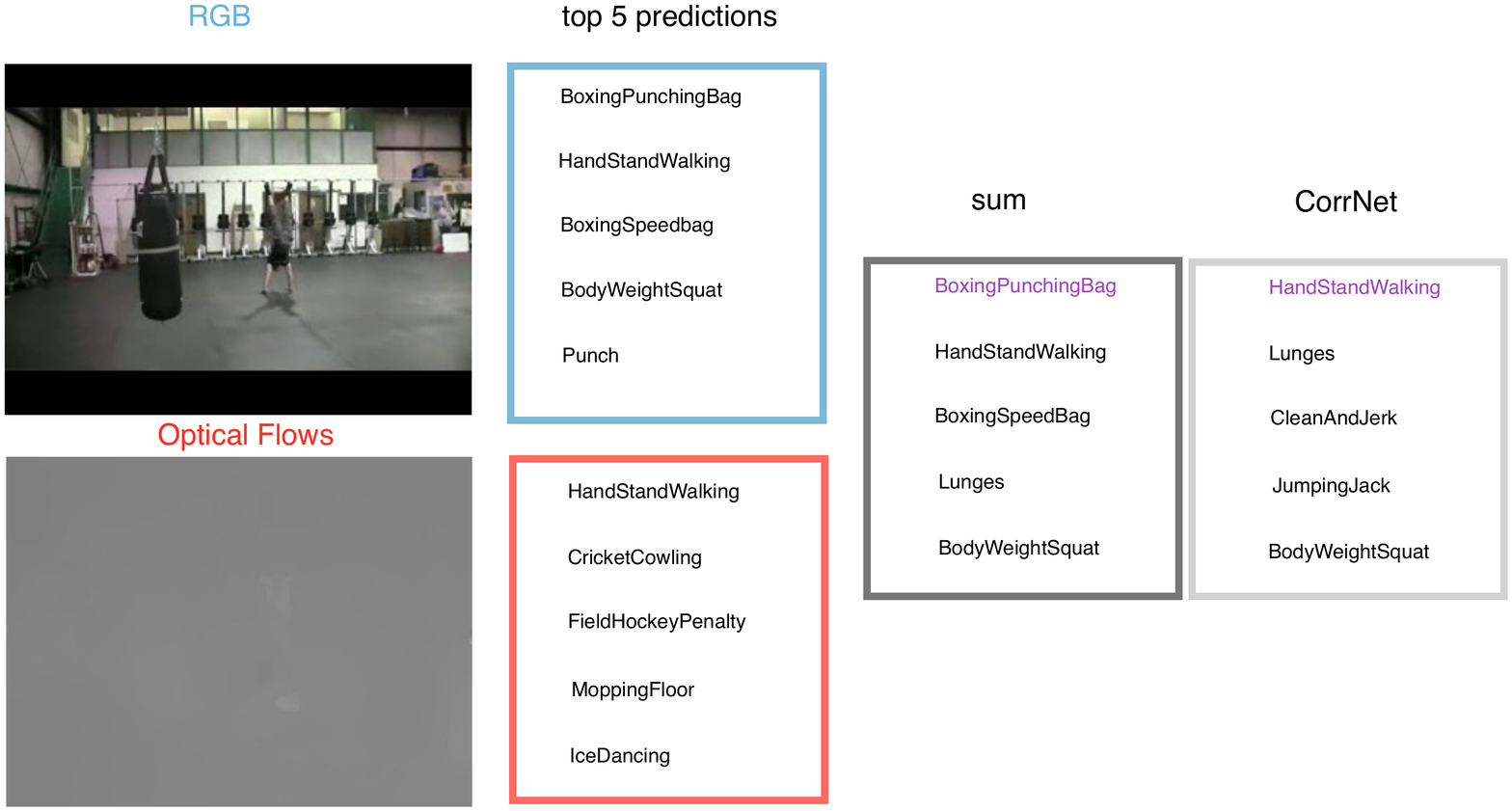}
  \caption{}
  \label{fig:class_score2}
\end{subfigure}
\caption{Top 5 inference results of RGB, Flow, sum fusion, and Corrnet of (a) Playing Cello and (b) Handstand walking video.}

\end{figure}

\subsection{Correlation Interpretation}

Fig. \ref{fig:class_score1} shows top 5 predictions from RGB and flow along with its sum and output of Corrnet for semantic comparison. Sum is class scores obtained by addition between RGB and flows stream, whereas Corrnet is obtained by correlation network. RGB stream infers \textit{playing violin}, \textit{playing cello}, \textit{playing flute}, \textit{archery}, and \textit{hulahoop} that influenced by shapes or textures. Flow stream predicts \textit{archery}, \textit{playing cello}, \textit{golf swing}, \textit{throw discuss}, and \textit{taichi} that influenced by motions. Given such prediction classes, Corrnet correctly infers \textit{playing cello} rather than playing violin which predicted by sum. Top 5 composition of Corrnet prediction also seems more reasonable in sense that there is no  \textit{hulahoop} but rather  \textit{playing flute} which has similar hand motion. Fig. \ref{fig:class_score2} shows another example of top 5 prediction from handstand walking action. Among top 5 inferences, Corrnet correctly spots \textit{handstand walking}, \textit{lunge}, \textit{clean and jerk}, \textit{jumping jack}, and \textit{body weight squat} of which make sense since they are semantically similar action. Whereas, result of sum shows \textit{boxing punching bag} and \textit{boxing speed bag} which very different from actual \textit{hand stand walking}. It shows that class predictions are still highly influenced by \textit{boxing punch} object.

\subsection{Comparison with sequence based model of LSTM}

LSTM is long standing method for capturing temporal dynamics from sequence data like videos. Our TSN-Corrnet shares similarity with that of TS-LSTM experiment done by \cite{ma2019ts} in terms of network, data augmentation, and training and testing. Our baseline TSN without LSTM cell delivers result of 93.5 \%. TSN-LSTM is trained on two setting of using the first 10 seconds of videos and full videos on UCF101 dataset.

\begin{table}[H]
  \centering
  \begin{tabular}{|c|c|c|}
  \hline
    & accuracy & duration\\
    \hline
    TSN&93.5&1 frame\\
    \hline
    TSN-LSTM &93.7& 10 seconds  \\
    \hline
    TSN-LSTM &94.1& full videos  \\
    \hline
    TSN-Corrnet (Shannon fusion)&94.4&1 frame   \\
    \hline
  \end{tabular}
  \caption{Accuracy of TSN given duration of clips used for training}
  \label{tab:9}
\end{table}

As shown in Table \ref{tab:9}, LSTM requires sequence of the videos to see temporal information. The accuracy becomes better as LSTM captures more frames of each video (93.7 \% to 94.2 \%). Our proposed fusion method obtained 94.4\% by looking only 1 frame per segment to achieve performance of which slightly better than LSTM. This confirms that our proposed method is able to capture video information with better performance.

\section{Conclusion}
\label{conclusion}
We have presented correlation network that capture spatiotemporal correlation given arbitrary timestamps. State-of-the-art CNN training of video recognition, however, is done on frame-by-frame basis using spatial and motion streams. The final layers of already trained spatial and temporal network are correlated to form two-dimensional correlation tensor. This is then fed to the three layers of full connected layers for training. Predictions are formed by fusing the output of correlation network with that from spatial and temporal stream's output.
Experimental results show that this correlation network contribute to an increase in recognition accuracy, revealing the importance of spatiotemporal correlation for long range video recognition.

\section*{Acknowledgment}

The authors would like to thank KAKENHI project no. 16K00239 for funding the research.

\bibliography{mybibfile}

\end{document}